%% file: egpaper_final.tex
\documentclass[10pt,twocolumn,letterpaper]{article}

\usepackage{iccv}
\usepackage{times}
\usepackage{epsfig}
\usepackage{graphicx}
\usepackage{amsmath}
\usepackage{amssymb}

\usepackage{booktabs} 
\usepackage{epsfig}
\usepackage{graphicx}
\usepackage{amsmath}
\usepackage{amssymb}
\usepackage{bm}
\usepackage{xcolor}
\usepackage{color,array}
\usepackage{multirow,makecell}
\usepackage{subfig}
\usepackage{float}
\usepackage[compatible]{algpseudocode}
\usepackage{algorithm}

\iccvfinalcopy

\ificcvfinal\fi
\begin{document}


\title{Multiple Kernel Fisher Discriminant Metric Learning for Person Re-identification}

\author{T M Feroz Ali \hspace{1cm} Kalpesh K Patel \hspace{1cm} Rajbabu Velmurugan \hspace{1cm} Subhasis Chaudhuri\\
Dept. of Electrical Engineering, Indian Institute of Technology Bombay\\
{\tt\small \{ferozalitm,kkp,rajbabu,sc\}@ee.iitb.ac.in}
}

%
%
%


\maketitle

\begin{abstract}
Person re-identification addresses the problem of matching pedestrian images across disjoint camera views. Design of feature descriptor and distance metric learning are the two fundamental tasks in person re-identification. In this paper, we propose a metric learning framework for person re-identification, where the discriminative metric space is learned using Kernel Fisher Discriminant Analysis (KFDA), to simultaneously maximize the inter-class variance as well as minimize the intra-class variance. 
We derive a Mahalanobis metric induced by KFDA and argue that KFDA is efficient to be applied for metric learning in person re-identification. 
We also show how the efficiency of KFDA in metric learning  can be further enhanced for person re-identification by using two simple yet efficient multiple kernel learning methods. We conduct extensive experiments on three benchmark datasets for person re-identification and demonstrate that the proposed approaches have competitive performance with state-of-the-art methods. 
\end{abstract}



\input{body_icvgip3_arxiv}


{\small
\bibliographystyle{ieee}
\bibliography{my_bibliography}
}

\end{document}

%% file: body_icvgip3_arxiv.tex
\section{Introduction}
\begin{figure*}[t]
\begin{minipage}{1\linewidth}
  \centering
  \centerline{\includegraphics[width=15cm]{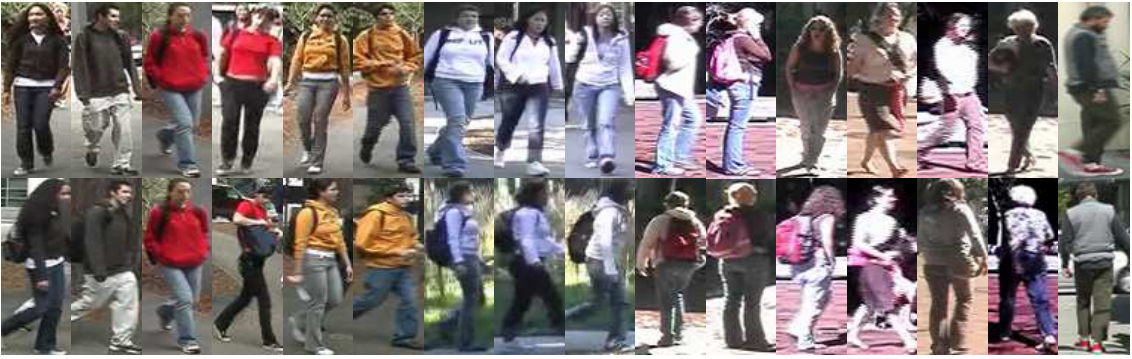}}
\end{minipage}
\caption{ Sample images from person re-identification dataset VIPeR, showing low resolution with large variations in illumination, viewpoint, background and pose. Images in each column correspond to the same person viewed in different cameras.}
\label{fig:re-ID}
\end{figure*}

Person re-identification (re-ID) addresses the problem of matching pedestrian images across disjoint camera views. Person re-ID is of growing interest due to its wide application in video surveillance, security, bio-metrics and forensics. It is a very challenging problem since images of the same person in distinct cameras look very different due to the large variations in illumination, pose, viewpoint, camera characteristics and background clutter (see Fig. \ref{fig:re-ID}). The low resolution makes the image quality insufficient to distinguish the identities based on their physical attributes. Moreover, the costume of distinct pedestrians can be very similar making them more indistinguishable.

Person re-ID consists of generally two stages: design of robust feature descriptors and distance metric learning. The feature descriptors are designed to capture the similarity of same person images and the dissimilarity of distinct persons. Various feature descriptors ELF \cite{ELF}, SDALF \cite{SDALF}, WHOS \cite{LisantiPAMI14}, LOMO \cite{LOMO}, GOG \cite{GOG} 
have been proposed in the literature. However, due to large variations in the camera and scene characteristics, the accuracy of the feature descriptors in discrimination is limited. Therefore distance metric learning  \cite{PRDC,LFDA:CVPR,SSSVM,rPcca,KISSME,LOMO,MLAPG,Zheng:nfst,NK3ML}
 is used as a second stage in person re-ID systems  to improve the discrimination. They learn a distance function which minimizes the within class variance and maximizes the between class variance to effectively improve the accuracy of person re-ID systems.

There exist re-ranking methods \cite{Symantic,song:scalableManifold,SHaPE,Re-ranking:k-reciprocal} to be applied in the testing phase of a trained person re-ID system, to further improve the accuracy. They refine the ranked results by learning the similarity or manifold structure of the whole test data. However, they are computationally intensive and unsuitable for real-time applications. Moreover, the accuracy of re-ranking methods largely depend on the efficiency of the preceding metric learning method. Hence, efficient metric learning methods are very crucial for such person re-ID systems also.
 \color{black}Deep learning has also been applied for person re-ID and shown good results on large training datasets \cite{Beyond:triplet_loss, Zhun:BodyLatentParts, Zhou:PointtoSetSimilarity, SpindleNet}. However, the availability of large labeled data for every new camera network is practically very expensive in the context of person re-ID. Regardless, metric learning methods have shown impressive performance even with small training data.   Therefore, developing efficient metric learning methods is very important for person re-ID and hence is the main focus of this paper.

In this paper, we propose a metric learning framework for person re-ID, where the discriminative metric space is learned using Kernel Fisher Discriminant Analysis (KFDA) \cite{kfda_Mika}. We derive a Mahalanobis distance metric, induced by KFDA, which simultaneously maximizes the inter-class variance as well as minimizes the intra-class variance. We first argue that, without the need of any supplementary method, KFDA  has strong potential and is \color{black}efficient to be applied for metric learning in person re-ID. In order to further enhance the efficiency of the learned metric in discrimination, we also propose to use two simple yet efficient multiple kernel learning frameworks. Multiple kernel learning is used to remove the bias in using a single kernel and to utilize the information from all the given kernels effectively.
We use kernels for two main reasons: (i) The training computation in most metric learning methods largely depends on the dimensionality of the feature descriptor, which is very high, typically in ten thousands. To remove the computational overhead, most methods use unsupervised dimensionality reduction, which removes the important discriminative information. However, use of kernels help reduce the computations to the order of the number of samples, without the need of any dimensionality reduction and thus providing significant reduction in the training time without losing discriminative information. 
(ii) Usage of appropriate kernel helps to get a richer representation of the data in the high dimensional \textit{feature space} for better discrimination. Kernels facilitate to model the high non-linearity in the person appearance across cameras for more efficient person re-ID. 

Our main contributions in this paper are the following: 
\begin{itemize}
\item We first identify that, without the need of any supplementary method, KFDA is a strong candidate to be applied in metric learning for person re-ID.
\item We show that the performance of KFDA for person re-ID can be further enhanced using two multiple kernel learning frameworks. 
\item We perform extensive experiments on three datasets and show that the proposed frameworks attain competitive performance with many state-of-the-art methods.
\end{itemize} 

\section{Related Methods}
The objective of metric learning methods is to learn a metric in which distances between the similar class samples are smaller compared to the dissimilar class samples.
 KISSME \cite{KISSME} used PCA and a log likelihood ratio test to derive a Mahalanobis distance metric. In PRDC \cite{PRDC}, the probability of a positive pair to have a smaller distance is maximized compared to a negative pair. Hirzer \textit{et. al} \cite{RPLM} relaxed the positive semi-definite constraint in the Mahalanobis metric learning. LADF \cite{LADF} generalized the metric learning using a locally adaptive decision rule. KNFST \cite{Zheng:nfst} learned a discriminative nullspace where all the samples of the same class are maximally collapsed to singular points by making the within class scatter to zero. 

The methods most related to our approach are LFDA \cite{LFDA:CVPR}, kLFDA \cite{rPcca}, MKML \cite{MKML} and XQDA \cite{LOMO}. Pedagadi \textit{et. al}  \cite{LFDA:CVPR} used \textit{Local} Fisher Discriminant Analysis (LFDA) \cite{LFDA:ICML} to learn the \textit{local} structure of multi-modal data to appropriately embed it using Locality Preserving Projection (LPP). Xiong \textit{et. al} \cite{rPcca} applied kernelized version of the LFDA, namely Kernel \textit{Local} Fisher Discriminant Analysis (kLFDA) for person re-ID. Though both LFDA and kLFDA have been developed in the framework of Fisher Discriminant Analysis (FDA), their primary focus has been to use the \text{local neighborhood} information of the data for metric learning. On the contrary, our approach does not utilize any such local information and still achieves higher accuracy than both LFDA and kLFDA. 
We infer that using the local neighborhood information as in LFDA and kLFDA, is not beneficial for person re-ID, especially for the small datasets, since there are very few samples per class to efficiently estimate local neighborhood information. Our work primarily focuses on the potential of KFDA as a metric learning method for person re-ID, without using any supplementary information like local neighborhood structure. To the best of our knowledge, no previous work exists that explore the direct applicability of KFDA individually in metric learning for person re-ID. In addition, we further enhance the efficiency of the KFDA based Mahalanobis metric using multiple kernel learning framework to effectively utilize the information from multiple kernels.

Another closely related work is Multiple Kernel Metric Learning (MKML) \cite{MKML}. It uses an exponential error function and an adaptive weighted kernel combination to iteratively learn an LFDA based distance metric for person re-ID, and is fundamentally different from our approach.
  The Cross-view Quadratic Discriminant Analysis (XQDA) \cite{LOMO} used the framework of Quadratic Discriminant Analysis (QDA) with zero mean Guassians to derive the KISSME metric and learn a subspace simultaneously. For subspace learning, the Fisher Discriminant Analysis (FDA) was not applicable due to zero mean of the class distributions and therefore the authors maximized the ratio of similar class and dissimilar class variance using an optimization technique motivated from FDA. Therefore XQDA is also primarily different from our approach. 

\section{Person Re-identification using Metric Learning}
\label{sec:PRUML}
In this section, we describe the framework for person re-ID using the proposed metric learning approaches. We first obtain a Mahalanobis distance metric induced by KFDA and show how this metric can be further improved using two different multiple kernel methods. Next we describe the person re-identification framework that uses the proposed metric learning approaches.

\subsection{Kernel Fisher Discriminant Analysis}
\label{ssec:KFDA}
Metric learning methods in person re-ID have shown to highly benefit from kernelization \cite{rPcca}. This is because of the high non-linearity in the appearance of the person  across the non overlapping cameras. Kernel methods use a non-linear function $\Phi(x)$ to map the input features to a high dimensional \textit{feature space} $\mathcal{F}$, where the classes are well separated. 

For a given set of $n$ samples $x_1, x_2,\ldots,x_n \in \mathbb{R}^{d}$ belonging to $c$ classes, the objective of Kernel Fisher Discriminant Analysis (KFDA) is to learn a discriminative subspace where similar class samples are closer and dissimilar class samples are well separated. KFDA calculates linear discriminants in the feature space $\mathcal{F}$ and thereby learns non-linear discriminants in the input data space.   The discriminant vectors $w_1,\ldots,w_p \in \mathbb{R}^{d}$ form the columns of the projection  matrix $W \in \mathbb{R}^{d \times p}$. Since each discriminant vector $w_k$ lies in the span of the mapped data $\Phi(x_1), \ldots, \Phi(x_n)$, we can express them as
\begin{equation}
\label{eqn:alpha}
w_k = \sum \limits_{l=1}^{n} (\alpha_{k})_l \Phi(x_l) \,,
\end{equation}
where $(\alpha_{k})_l$ is the $l$th element of the expansion vector $\alpha_k \in \mathbb{R}^{n}$. The discriminant vectors of KFDA are estimated such that they maximize the \textit{Fisher criterion}, i.e.,
\begin{equation}
\label{eqn:EqKMMC}
\begin{aligned}
& \underset{W}{\text{maximize}} & &  tr\{(W^T S_w^{\Phi} W)^{-1}(W^T S_b^{\Phi} W)\}\,,
\end{aligned}
\end{equation}
where, $tr$ represents trace, $S_b^{\Phi}$ is the between class scatter matrix and $S_w^{\Phi}$ the within class scatter matrix in the feature space.
\begin{eqnarray}
\begin{aligned}
S_w^{\Phi}  &= \sum_{i = 1}^{c} \sum_{j=1}^{n_i} (\Phi(x_j^{(i)})-m_i^{\Phi})(\Phi(x_j^{(i)})-m_i^{\Phi})^T \\
S_b^{\Phi}  &=\sum_{i = 1}^{c} n_i(m_i^{\Phi}-m^{\Phi})(m_i^{\Phi}-m^{\Phi})^T
\end{aligned}
\end{eqnarray}
$n_i$ is number of samples in class $i$, $m_i$ the mean of class $i$ and $m$ the mean of all samples. Equivalently, KFDA learns a discriminative subspace to maximize the between class variance and minimize the within class variance.

The problem in (\ref{eqn:EqKMMC}) can be formulated in terms of inner products $\langle \Phi(x_i), \Phi(x_j) \rangle $ of the input data and is replaced by an appropriate positive definite kernel function $ k(x_i,x_j)$. As the kernel function can be easily computed, there is no need to explicitly calculate the non-linear function $\Phi(x)$. 

Using (\ref{eqn:alpha}) along with the definitions $m_i^{\Phi} = \frac{1}{n_i} \sum_{j=1}^{n_i}  \Phi(x)$ , $m^{\Phi} =  \frac{1}{n} \sum_{i=1}^{c}n_i m_i^{\Phi}$ and later replacing the inner product using the kernel function, we get $w_k^T m^{\Phi} =  \alpha_{k}^T \widetilde{m} $ and
$w_k^T m_i^{\Phi} = \alpha_{k}^T \widetilde{m}_i$
where $\widetilde{m} := \frac{1}{n} \sum_{i=1}^{c} n_i \widetilde{m}_i$ and $(\widetilde{m}_i)_l := \frac{1}{n_i}  \sum_{j=1}^{n_i}    k(x_l,x_j^{(i)})$. 
The between class and within class scatter can be, respectively, rewritten as $w_k^T S_b^{\Phi} w_k = \alpha_{k}^T P \alpha_{k}$ and $w_k^T S_w^{\Phi} w_k = \alpha_{k}^T Q \alpha_{k}$ where  
\begin{eqnarray}
\begin{aligned}
P &=\sum_{i = 1}^{c} n_i (\widetilde{m}_i-\widetilde{m})(\widetilde{m}_i- \widetilde{m})^T,\\
Q &= \sum_{i = 1}^{c}  K_i(I_{n_i}- \frac{1}{n_i} 1_{n_i}1_{n_i}^T)K_i^T \,,
\end{aligned}
\end{eqnarray}
$K_i$ is ${(n \times n_i)}$ kernel matrix of the $i$th class defined as $(K_i)_{uv}:=k(x_u,x_v^{(i)})$, $I_{n_i}$ is $(n_i \times n_i)$ identity matrix and $1_{n_i}$ is $n_i$ dimensional vector of ones.
The final optimization problem of KFDA becomes:
\begin{equation}
\label{eqn:EqKMMC2}
\begin{aligned}
& \underset{A}{\text{maximize}} & &  tr\{(A^T Q A)^{-1}(A^T P A)\} \,,
\end{aligned}
\end{equation}
where $A = [\alpha_1,\ldots,\alpha_p]$. If $Q$ is singular, the optimal discriminants of KFDA are obtained using the first $p$ leading eigenvectors of $Q^{-1}P$.

\subsection{Metric Learning using KFDA}
Kernel Local Fisher Discriminant Analysis (kLFDA) \cite{rPcca} learns a distance metric utilizing the  local neighborhood information of the data. They learn a discriminative space to appropriately embed the structure of multi-modal data using Locality Preserving Projection (LPP), using the framework of KFDA. However, most typical datasets in person re-ID are of small size with very few samples per class. This could limit the proper estimation of the local neighborhood structure. Hence in our approach, we first demonstrate how to obtain a Mahalanobis distance metric directly induced by KFDA, without using any supplementary techniques like local neighborhood estimation as in LFDA, kLFDA and MKML. \color{black}

A Mahalanobis matrix $M \in \mathbb{R}^{d \times d}$ is a positive semi-definite matrix and hence can be decomposed into the form $M=VV^T$, where $V \in \mathbb{R}^{d \times p}$. In our approach, we directly choose $V$ to be the projection matrix $W$ of KFDA, since such a learned Mahalanobis distance metric between any two samples $\Phi(x_i)$ and $\Phi(x_j)$ in the feature space $\mathcal{F}$ has equivalence with the Euclidean distance between the samples in the discriminative subspace of KFDA. The equivalence can be seen by the below analysis.
\begin{eqnarray}
&&\hspace{-1.5cm}(\Phi(x_i)-\Phi(x_j))^T M (\Phi(x_i)-\Phi(x_j))\nonumber \\ 
&=& (\Phi(x_i)-\Phi(x_j))^T VV^T (\Phi(x_i)-\Phi(x_j))\\ 
&=& (\Phi(x_i)-\Phi(x_i))^T WW^T (\Phi(x_i)-\Phi(x_j))\\ 
&=& \Vert W^T\Phi(x_i) - W^T\Phi(x_j) \Vert^2
\label{eqn:MahKFDA1}
\end{eqnarray}
Any general feature vector $\Phi(y)$ in the feature space $\mathcal{F}$ can be projected onto the discriminant vector $w_k$ as follows:
\begin{eqnarray}
w_k^T \Phi(y) &= \Big( \sum\limits_{l = 1}^{n} (\alpha_{k})_l \Phi(x_l)\Big)^T \Phi(y) \\
&= \sum\limits_{l = 1}^{n} (\alpha_{k})_l \langle (\Phi(x_l), \Phi(y) \rangle 
&= \alpha_{k}^T k_{y}\,,
\label{eqn:MahKFDA2}
\end{eqnarray}
where $k_{y} = [k(x_1,y),\ldots,k(x_n,y)]^T$. Using (\ref{eqn:MahKFDA1}) and (\ref{eqn:MahKFDA2}), we solve the Mahalanobis distance metric as:
\begin{eqnarray}
&&\hspace{-1 cm}(\Phi(x_i)-\Phi(x_j))^T M (\Phi(x_i)-\Phi(x_j))\nonumber \\ 
&=& \Vert W^T\Phi(x_i) - W^T\Phi(x_j) \Vert^2 \nonumber\\
&=& \Vert [w_1, w_2, \ldots, w_p ]^T\Phi(x_i) - [w_1, w_2, \ldots, w_p ]^T\Phi(x_j) \Vert^2 \nonumber\\
&=& \Vert [\alpha_1, \alpha_2, \ldots, \alpha_p ]^T k_{x_i} - [\alpha_1, \alpha_2, \ldots, \alpha_p ]^Tk_{x_j} \Vert^2 \nonumber\\
&=& \Vert A^T (k_{x_i} - k_{x_j}) \Vert^2 \,.
\label{eqn:MahKFDA3}
\end{eqnarray}
Thus the Mahalanobis distance metric is obtained directly using the KFDA, without any supplementary techniques. It has a closed form solution with very  few parameters and is efficient for practical implementation. We next propose an improvement to  this framework using multiple kernel learning. 

\subsection{Multiple Kernel Fisher Metric Learning}
Multiple kernel learning (MKL) is an efficient way of enhancing the accuracy of  kernel based methods. Instead of using a single kernel, MKL methods find an optimal combination of a given set of pre-specified kernels. Using multiple kernels help to remove the possible bias in using a single kernel and finding a better solution. The MKL methods generally use a linear or non-linear combination of the given kernels. A linear combination of kernels is equivalent to scaling the feature space corresponding to each kernel and concatenating them as a single feature representation. 
Optimization based methods are the most popular for MKL using KFDA. However, empirically we find that for person re-ID such methods are unstable and their performance gets degraded for small size datasets, where there are very few samples in each class.

\begin{table*}[t]
\vspace{-10pt}
\small
\caption{Person re-ID accuracy (in \%) comparison with baseline methods using (a) GOG and (b) LOMO feature descriptors on the GRID dataset. The proposed methods are shown in bold.}
\begin{center}
\subfloat[Subtable 1 list of tables text][GOG]{
\resizebox{0.99\columnwidth}{!}{%
\begin{tabular}{lccccc}
\hline 
Methods & Rank-1 & Rank-5 & Rank-10 & Rank-15 & Rank-20 \\
\hline
KISSME\cite{KISSME} &21.36 & 	40.64 & 	51.20 & 	57.44 & 	62.56\\
LFDA\cite{LFDA:CVPR} &21.60	 & 39.60	 & 51.68 & 	58.32	 & 63.28\\
XQDA\cite{LOMO} & 24.80	 & 46.96	 & 58.40 & 	64.00	 & 68.88\\
kLFDA\cite{rPcca} &23.76	 & 45.60 & 	53.44	 & 59.92 & 	64.48\\
MFA\cite{rPcca} &24.16	 & 44.08 & 	55.76 & 	61.28 & 	65.12\\
KNFST\cite{Zheng:nfst} &24.88 & 	43.68 & 	53.28 & 	61.04 & 	65.44\\
\hline
\textbf{KFDA}  & 24.96 & 	44.24 & 	54.80 & 	61.60 & 	66.24 \\
\textbf{NP-MFML} & \color{red}25.76 & 	\color{red}48.56	 & \color{red}60.24	 & \color{red}67.20 & 	\color{red}70.80\\
\textbf{SM-MFML} & \color{blue}25.04	 & \color{blue}46.40 & 	\color{blue}57.76 & 	\color{blue}63.36 & 	\color{blue}68.80\\
\hline
\end{tabular}
\label{table:baseline1}
}
}
\quad 
\subfloat[Subtable 2 list of tables text][LOMO]{
\resizebox{0.99\columnwidth}{!}{%
\begin{tabular}{lccccc}
\hline
Methods & Rank-1 & Rank-5 & Rank-10 & Rank-15 & Rank-20 \\
\hline
KISSME\cite{KISSME} &11.12	&26.80	&36.32	&43.84	&48.96\\
LFDA\cite{LFDA:CVPR} &13.28	&29.20	&36.88&	42.72&	46.88\\
XQDA\cite{LOMO} & 16.56&	33.84	&41.84	&47.68	&52.40\\
kLFDA\cite{rPcca} &14.24	&28.40	&37.52&	43.84	&49.36\\
MFA\cite{rPcca} &15.12	&30.80	&40.80	&46.48	&52.40\\
KNFST\cite{Zheng:nfst}  &14.88&	29.76	&41.28	&46.56&	50.88\\
\hline
\textbf{KFDA}  &14.32	&29.68&	38.88	&43.92	&48.56\\
\textbf{NP-MFML} & \color{red}18.32&	\color{blue}36.24	&\color{red}47.76	&\color{red}54.40	&\color{red} 59.68\\
\textbf{SM-MFML} & \color{blue}18.00	&\color{red}36.72	& \color{blue}46.32&	\color{blue}53.04	&\color{blue}57.92\\
\hline
\end{tabular}
\label{table:baseline2}
}
}
\end{center}
\label{table:baseline}
\vspace{-3mm}
\end{table*}

In our approach, we propose to use two simple yet efficient multiple kernel learning methods for metric learning using KFDA. The first method is adapted from Proportionally Weighted Multiple Kernels (PWMK) \cite{Tanebe}. PWMK seeks to learn a convex combination of $q$ pre-defined kernels, based on the relative importance of the individual base kernels (estimated using cross validation). 
\begin{equation}
 \widetilde{k}(x_i,x_j) = \sum_{t=1}^q \beta_t k_t(x_i,x_j) \quad s.t \quad \sum_{t=1}^q \beta_t = 1
 \label{eg:MKL1}
\end{equation}
$\beta_1,\ldots,\beta_q$ are the weights used for the convex combination of the base kernels $k_1,\ldots,k_q$. The PWMK selects the kernel weights based on the following rule:
\begin{equation}
\beta_t =  \frac{\pi_t-\delta}{\sum_{r=1}^q (\pi_r - \delta)} \,,
\end{equation}
where $\pi_r$ is the accuracy obtained using the $r$th kernel. The value of the threshold $\delta$ is chosen less than or equal to the minimum accuracy obtained among all the individual kernels.
However, such an approach is empirically seen to assign significant weights to almost all the kernels including the poorly performing kernels. In-order to suppress the bias of the poorly performing kernels, we learn the kernel weights based on the following modified function.
\begin{equation}
    \beta_t = 
\begin{cases}
    \frac{\pi_t-\delta^{'}}{\sum_{r\in \mathcal{S}} (\pi_r - \delta^{'})},& \text{if } t\in \mathcal{S}\\
    0,              & \text{otherwise}
\end{cases}
\end{equation}
The threshold $\delta^{'}$ is chosen to be the accuracy of $(N+1)$th best performing kernel and the set $\mathcal{S}$ consists of the $N$ best performing kernels. This approach assigns null weight to all the kernels except the top $N$ best performing kernels, and thus removes the contribution of poorly performing kernels in learning the final kernel. This also ensures that the learned  kernel weights are non-negative, ensuring positive semi-definiteness of the resultant kernel and also satisfies the constraint in (\ref{eg:MKL1}). We use the above approach of multiple kernel learning along with our KFDA based Mahalanobis metric learning and refer the resulting framework as \textit{\textbf{N}-\textbf{P}roportionally Weighted \textbf{M}ultiple Kernel \textbf{F}isher \textbf{M}etric \textbf{L}earning (NP-MFML)}.

The second approach for multiple kernel learning used in our framework 
is based on the \textit{Squared Matrix} (SM) method, which was earlier proposed for SVM in \cite{Diego1,Diego2}. Given two kernel matrices $K1$ and $K2$, the method builds the final kernel matrix $\widetilde{K}$ using the below form
\begin{equation}
\widetilde{K} = \frac{1}{2}(K_1+K_2) + \tau f(K_1-K_2) \,.
\label{eqn:SM_MFML}
\end{equation}%
The  term $f(K_1-K_2)$ is chosen so that it quantifies the difference in information that the kernel matrices $K1$ and $K2$ provide for classification. The difference term should vanish when $K1$ and $K2$ provide similar classification results, yielding $\widetilde{K} \simeq K1 \simeq K2$. The scaling factor $\tau$ is a positive constant that controls the relative importance of the  difference function. In our multiple kernel framework, we set the difference function to be:
\begin{equation}
f(K_1-K_2) = (K_1-K_2)(K_1-K_2) \,.
\end{equation}
This square matrix form is chosen since it guarantees positive semi-definiteness of the final kernel matrix $\widetilde{K}$, making it a valid Mercer kernel. We refer this approach of multiple kernel learning with our   KFDA induced metric learning as \textit{\textbf{S}quare \textbf{M}atrix-\textbf{M}ultiple Kernel \textbf{F}isher \textbf{M}etric \textbf{L}earning (SM-MFML)}. 

\subsection{Person Re-identification}
Training in a person re-ID system consists of two stages. In the first stage, we extract feature descriptors (such as GOG \cite{GOG} or LOMO \cite{LOMO}) for all the images in the training set. In the second stage, we learn the distance metrics (to better capture the similarity of same person images and the dissimilarity of distinct persons) and the corresponding transformations, as discussed above. During testing, the feature descriptors are obtained for all the images in the testing set and the appropriate transformations are applied. The test set is divided into probe set and gallery set. Each  probe is matched against the complete gallery. The matching scores are calculated using the distance metric. The gallery images are sorted based on the matching scores and then finally ranked to indicate their closeness to the probe image. 

Let $x_1,\ldots,x_n$ be the feature descriptors generated for the given $n$  training images belonging to the $c$ classes. For training using KFDA, we choose a kernel $k$ and solve the optimization problem 
(\ref{eqn:EqKMMC2}), to obtain $A$. Let $y$ and $z$ be the feature descriptors for a given test probe and a test gallery image, respectively. Their matching score in the discriminative subspace of KFDA can be calculated as:
\begin{eqnarray}
(\Phi(y)-\Phi(z))^T M (\Phi(y)-\Phi(z)) = \Vert {A}^T (k_{y} - k_{z}) \Vert^2 \,.
\label{eqn:MahKFDA4}
\end{eqnarray}
A lower (higher) score corresponds to the closeness (difference) between the probe $y$ and the gallery image $z$. The matching scores are sorted and the corresponding gallery images are ranked to obtain the most relevant matches to the probe image. This completes the process of person re-identification.

For the multiple kernel frameworks NP-MFML and SM-MFML, we use the corresponding resultant kernel $\widetilde{k}$, obtained from the $q$ predefined kernels, for solving the optimization problem (\ref{eqn:EqKMMC2}) and obtain the solution $\widetilde{A}$. The remaining procedure is similar to that of the single kernel framework.

\section{Experimental Results}
\label{sec:Exp}
We conducted our experiments on various datasets using the framework discussed in Section \ref{sec:PRUML}. We next describe the datasets, performance measures and the parameter settings used to evaluate the proposed methods. We also compare them with several state-of-the-art methods.\\ 

\setlength{\parskip}{-0.75em}
\noindent\textbf{Datasets}: We evaluate our proposed method on three benchmark datasets: PRID450S \cite{PRID450S},  GRID \cite{GRID1} and VIPeR \cite{ELF}. They have 450, 250, 632 persons observed from two non overlapping cameras, respectively. All datasets have one image from each camera. The testing identities for person re-ID are generally considered unseen during training. Therefore, following the standard set up, the identities are divided equally for training and testing set, each having disjoint identities.  The test set images are equally divided into probe and gallery set. The test set of the GRID dataset has another 775 images in the gallery, whose identities are different from the 250 identities. \\

\noindent\textbf{Performance measure}: For testing, each probe image is matched against all the gallery images. The corresponding scores are sorted and the rank-K accuracy is calculated based on the occurrence probability of a true match within the top K ranks. The procedure is repeated ten times and the average scores are reported.\\

\noindent\textbf{Visual features}: We use GOG\cite{GOG} as the default feature descriptor in our method. GOG has a dimension of 27,622. It is extracted by hierarchically modeling  color and texture information as multiple Guassian distributions.\\

\noindent\textbf{Parameter settings:} 
We use RBF kernels in all our methods. For KFDA (single kernel), the kernel width is automatically determined using the root mean squared pairwise distance among the samples following \cite{Zheng:nfst,NK3ML}. For MFML, we use $q=20$ RBF kernels whose kernel widths are selected from the range 0.1 to 10 in log scale. The performance of each of these base kernels is estimated using a ten fold cross validation on the training data. The best two performing kernels are used in SM-MFML. The parameter $\tau$ of SM-MFML and $N$ of NP-MFML are also selected using cross validation.

\setlength{\parskip}{0em}
The matrix $Q$ that needs to be inverted for solving the optimization problem (8) is of size $n \times n$. However, the matrix $Q$ can have a rank of at most $n-c$, making it non-invertible. Hence we add a small regularizer to the diagonal elements of $Q$. We empirically find that a value of $10^{-7}$ as the regularizer is generally useful. 
For KFDA, the maximum number of discriminant vectors available is $c-1$. In our approach, we set the number of discriminant vectors $p$ to be $c-1$ to accommodate all the useful discriminative information. \\

\begin{figure*}[ht]
\begin{tabular}{ccc}
\includegraphics[width=0.9\columnwidth]{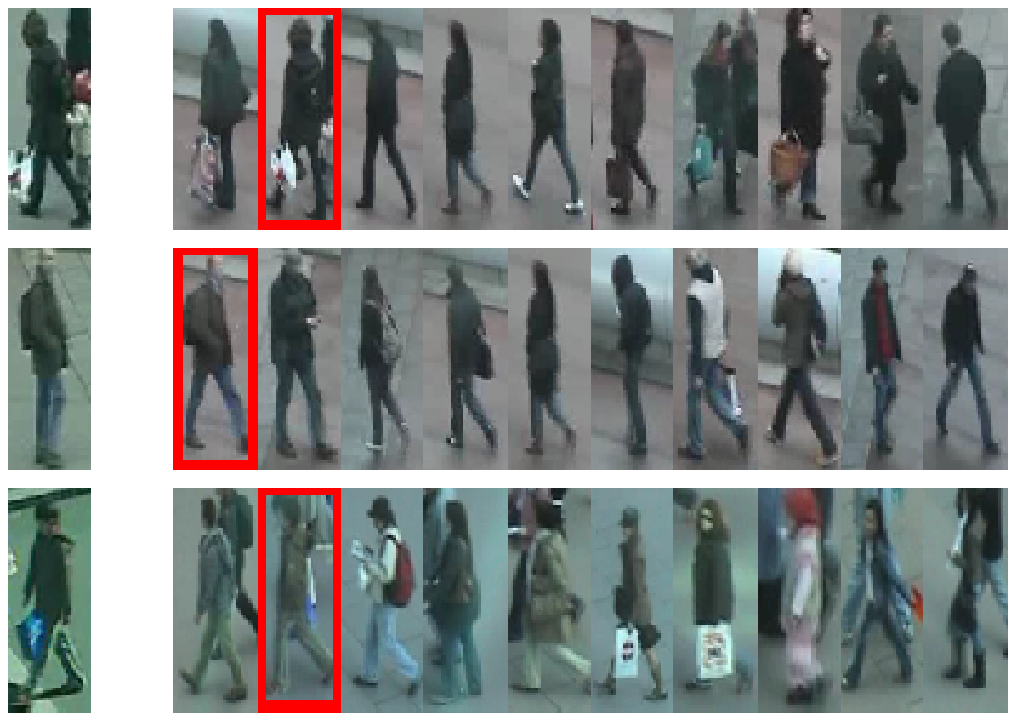} & \hspace{4em}
\includegraphics[width=0.9\columnwidth]{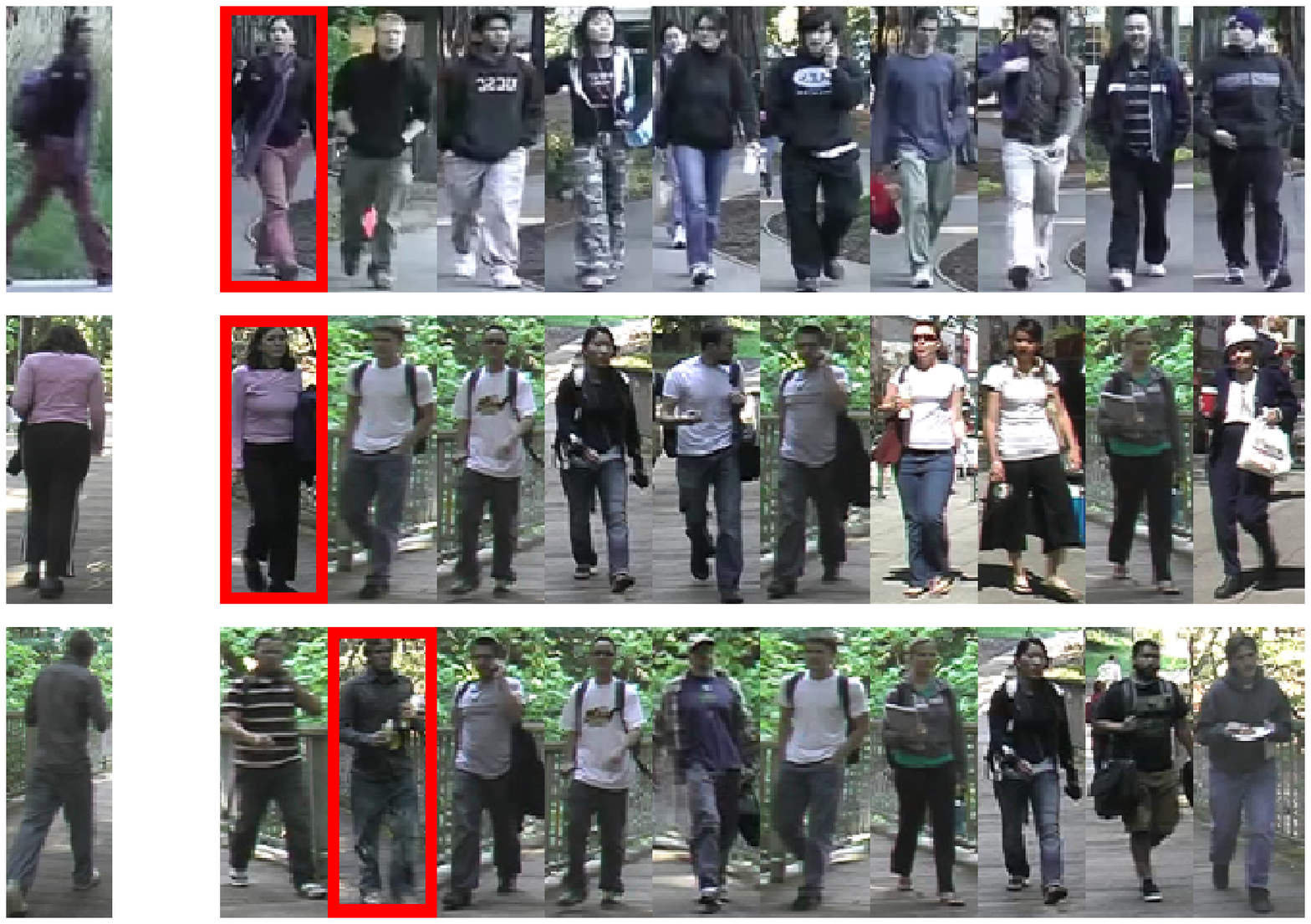}\\
(a)&(b)
\end{tabular}
\caption{ Illustration of probe images (left column) and their ranked results from the gallery using (a) SM-MFML on PRID450S dataset and (b) NP-MFML on VIPeR dataset. Sequence is from left to right. True match is shown with red border.}
\label{fig:teaser}
\end{figure*}

\subsection{Comparison with baselines}
We compare the performance of the proposed approaches KFDA, NP-MFML and SM-MFML with the baseline metric learning methods, whose codes are publicly available. They include KISSME \cite{KISSME}, LFDA \cite{LFDA:CVPR}, XQDA \cite{LOMO}, kLFDA \cite{rPcca}, MFA \cite{rPcca} and KNFST \cite{Zheng:nfst}. For fair comparison, we evaluate these methods using the same set of feature descriptors. We use GOG\cite{GOG} and LOMO\cite{LOMO} descriptors separately to evaluate their performance on GRID dataset and the ranking results are reported in Table \ref{table:baseline1} and \ref{table:baseline2} respectively. For the GOG descriptor, KFDA outperforms KISSME, LFDA, XQDA, kLFDA, MFA and KNFST by a margin of 3.60\%, 3.36\%, 0.16\%, 1.20\%, 0.80\% and 0.06\% respectively. This demonstrates that, without using any supplementary techniques, our Mahalanobis distance metric learned using KFDA is efficient and competitive with the baseline metric learning methods.  

It should be noted that KFDA outperforms both LFDA and kLFDA at almost all the ranks. They were designed to embed the local neighborhood information of the data using Local Binary Pattern (LBP) in the framework of FDA and KFDA respectively. However, our experimental results show that their performance is inferior compared to KFDA for person re-ID, indicating that there is no added advantage of learning the local structure of the data. One possible reason is that the typical datasets for person re-ID have very few samples per class, limiting  proper estimation of their local neighborhood information. 

Table \ref{table:baseline} also demonstrates that the proposed multiple kernel metric learning methods NP-MFML and SM-MFML  improves upon the KFDA based metric learning. In particular for LOMO descriptor, NP-MFML and SM-MFML boosts the rank-1 accuracy of KFDA by 4.00\% and 3.68\%,  respectively. This emphasize that our multiple kernel frameworks can effectively extract information from multiple kernels and remove the bias of using a single kernel. NP-MFML also attains an improvement margin of 7.20\%, 5.04\%, 1.76\%, 4.08\%, 3.2\% and 3.44\% against KISSME, LFDA, XQDA, kLFDA, MFA and KNFST, respectively. Similar improvements are seen using NP-MFML also for both LOMO and GOG descriptors. This particularly illustrates that the performance of the proposed methods is not because of a specific feature descriptor. \color{black}

\begin{table}[t]
\caption{Person re-ID accuracy (in \%) comparison with state-of-the-art results on GRID dataset.  Red and blue colors are used for the best and the second best scores respectively. * indicates re-ranking based methods.}
\begin{center}
\small
\resizebox{0.85\columnwidth}{!}{%
\begin{tabular}{lcccc}
\hline
Methods & Ref & Rank1 & Rank10 & Rank20 \\
\hline
\small MtMCML	&	\cite{MtMCML}	&	14.08	&	45.84	&	59.84	\\
PolyMap	&	\cite{ExPolyFeatMap}	&	16.30	&	46.00	&	57.60	\\
LOMO+XQDA	&	\cite{LOMO}	&	16.56	&	41.84	&	52.40	\\
MLAPG	&	\cite{MLAPG}	&	16.64	&	41.20	&	52.96	\\
KEPLER	&	\cite{KEPLER}	&	18.40	&	50.24	&	61.44	\\
DR-KISS	&	\cite{DR-KISS}	&	20.60	&	51.40	&	62.60	\\
KNFST  & \cite{Zheng:nfst} & 14.88	&	41.28	&	50.88 \\
SSSVM & \cite{SSSVM} & 22.40	& 51.28	&	61.20\\
SCSP	&	\cite{SCSP}	&	24.24	&	54.08	&	65.20	\\
GOG+XQDA	&	\cite{GOG}	&	24.80	&	\color{blue}58.40	&	\color{blue}68.88	\\
\textbf{NP-MFML} & Ours & \color{red}25.76	&	\color{red}60.24	&	\color{red}70.80\\
\textbf{SM-MFML}  & Ours  &\color{blue}25.04	&	57.76	&	68.80\\
\hline
*SSDAL	&	\cite{SSDAL}	&	22.40	&	48.00	&	58.40	\\
*SSM & \cite{song:scalableManifold} & \color{red}27.20	&	\color{red}61.12	&	\color{red}70.56\\
\hline
\end{tabular}
}
\end{center}
\label{table:GRIDall}
\vspace{-3mm}
\end{table}

\subsection{Comparison with state-of-the-art}
\noindent\textbf{Method of comparison}:
We compare the performance of our proposed frameworks with the existing state-of-the-art methods on all the three datasets. Apart from metric learning methods, there exists few re-ranking based methods proposed for person re-ID. Re-ranking is a post-processing stage applied over an existing metric learning method, for refining the ranked results using the entire training and test data.
Since our work focuses on metric learning, direct comparison of our results with such post-processing based methods is not advisable.  However, even with such a comparison, our proposed method attains competitive performance against the post-processing based methods. Moreover, our proposed metric learning approach is general enough to be integrated with any such post-processing based method to further enhance their accuracy. 
We list the results of such post-processing methods in separate rows for completeness. \\

\setlength{\parskip}{-0.75em}
\noindent\textbf{Experiments on GRID dataset}:
The images of GRID datasets are captured using 8 disjoint cameras installed in a busy underground station. It is a challenging dataset 
due to pose variations, lighting changes and low resolution images.
The existing state-of-the-art results on GRID dataset are reported in Table \ref{table:GRIDall}. NP-MFML outperforms all the metric learning methods. At rank-1, it attains an extra margin of 9.46\%, 9.12\%, 10.88\%, 3.36\%  and 0.96\% against PolyMap, MLAPG, KNFST, SSSVM and GOG+XQDA. Similar improvements are seen for SM-MFML also. It can also be observed that SM-MFML and NP-MFML performs competitively even with the re-ranking methods.  NP-MFML achieves a significant margin over SSDAL \cite{SSDAL}, and also performs very competitively with SSM \cite{song:scalableManifold}.\\

\begin{table}[t]
\caption{Performance comparison with state-of-the-art methods on PRID450S dataset.}
\small
\begin{center}
\resizebox{0.85\columnwidth}{!}{%
\begin{tabular}{lcccc}
\hline
Methods & Ref & Rank1 & Rank10 & Rank20 \\
\hline
SCNCD	&	\cite{colornames}	&	41.60	&	79.40	&	87.80	\\
LOMO+XQDA	&	\cite{LOMO}	&	59.78	&	90.09	&	95.29	\\
CSL	&	\cite{CSL}	&	44.40	&	82.20	&	89.80	\\
KNFST &	\cite{Zheng:nfst} & 59.47	&	91.96	&	96.53\\
SSSVM	&	\cite{SSSVM}	&	60.49	&	88.58	&	93.60	\\
GOG+XQDA	&	\cite{GOG}	&	\color{red}68.00 	&	\color{blue}94.36	&	97.64	\\
TMA	&	\cite{TMA}	&	52.89	&	85.78	&	93.33	\\
\textbf{NP-MFML} & Ours	&   \color{blue}67.38	&	\color{blue}94.36	&	\color{blue}97.82\\
\textbf{SM-MFML}  &Ours	&  67.29	&	\color{red}94.44	&	\color{red}97.78\\
\hline 
*Semantic	&	\cite{Symantic}	&	44.90	&	77.50	&	86.70	\\
*SSM	&	\cite{song:scalableManifold}	&	\color{red}72.98	&	\color{red}96.76	&	\color{red}99.11	\\
\hline
\end{tabular}
}
\end{center}
\label{table:PRID450Sall}
\end{table}

\noindent\textbf{Experiments on PRID450S dataset}:
In Table \ref{table:PRID450Sall}, we compare the performance of our proposed approaches with the state-of-the-art results on PRID450S dataset. Both NP-MFML and SM-MFML outperform almost all the state-of-the-art person re-ID methods. For example, NP-MFML attains a significant improvement margin of 14.49\% and 6.89\% against TMA and SSSVM, at rank-1. Though the proposed methods are slightly inferior to GOG+XQDA at rank-1, their results are better at rank-10 and rank-20. Also, the proposed methods significantly outperform the re-ranking method Semantic \cite{Symantic}. The ranking results using SM-MFML for few sample probe images are shown in Fig. \ref{fig:teaser} (a).\\

{
\begin{table}[t]
\caption{Comparison with state-of-the-art results on VIPeR dataset. `**' represents deep learning based methods.}
\begin{center}
\resizebox{0.85\columnwidth}{!}{%
\begin{tabular}{lcccc}
\hline
Methods & Ref & Rank1 & Rank10 & Rank20 \\
\hline
KISSME & \cite{KISSME}  & 19.60 &62.20 &77.00\\
LFDA & \cite{LFDA:CVPR}		&	24.18	&	67.12	&	-	\\
MLFL & \cite{midlevel}		&	29.11	&	65.95	&	79.87	\\
CAMEL & \cite{CAMEL}  & 30.90 &	-	&	-	\\
kLFDA & \cite{rPcca}  &32.30 & 79.70 & 90.90\\
PolyMap & \cite{ExPolyFeatMap}	&	36.80	&	83.70	&	91.70	\\
MKML & \cite{MKML}  &36.97 & 80.68 & 90.76\\
LOMO+XQDA & \cite{LOMO}		&	40.00	&	80.51	&	91.08	\\
MLAPG & \cite{MLAPG}		&	40.73	&	82.34	&	92.37	\\
\textit{l}1-graph & \cite{UlGraph}		&	41.50	&	-	&	-	\\
KNFST & \cite{Zheng:nfst}		&	42.28 &	82.94	&	92.06 \\
SSSVM & \cite{SSSVM}		&	42.66	&	84.27	&	91.93	\\
GOG+XQDA & \cite{GOG}		&	49.72	&	88.67	&	94.53	\\
**Shi et al. & \cite{Shi}		&	40.91	&	-	&	-	\\
**ImprovedDeep & \cite{ImprDeep}		&	34.81	&	-	&	-	\\
**S-CNN & \cite{SCNN}		&	37.80	&	66.90	&	-	\\
**DGD & \cite{DGD}		&	38.60	&	-	&	-	\\
**S-LSTM & \cite{SLSTM}		&	42.40	&	79.40	&	-	\\
**Quadruplet & \cite{Beyond:triplet_loss} & 49.05	&	81.96 & -	\\
\textbf{NP-MFML}  & Ours	&\color{red}50.76	&	\color{red}89.72	&	\color{blue}95.00\\ 
\textbf{SM-MFML}  &Ours	&\color{blue}50.47	&	\color{blue}89.30	&	\color{red}95.06\\
\hline
*Semantic & \cite{Symantic}	&	41.60	&	86.20	&	95.10	\\
*SSDAL & \cite{SSDAL}		&	43.50	&	81.50	&	89.00\\
*MuDeep & \cite{MuDeep}  & 43.03	&	85.76	& - \\
*OL-MANS & \cite{OnlineNegSamples}  & 44.97	&	84.97	&	93.64\\
*DLPAR & \cite{DLPAR}  & 48.70	&	85.10	&	93.00\\
*PDC & \cite{PDC}  & 51.30	&	84.20	&	91.50\\
*SSM & \cite{song:scalableManifold}		&	\color{red}53.70	&	\color{red}91.50	&	\color{red}96.10   \\
\hline
\end{tabular}
}
\end{center}
\label{table:viperall}
\end{table}
}

\noindent\textbf{Experiments on VIPeR dataset}:
VIPeR is one of the most challenging dataset for person re-ID as the images were captured from outdoor scene environment with large viewpoint changes, pose variation and significant illumination difference. It is the most popular evaluated dataset for person re-identification. The performance comparison of the proposed approaches on the VIPeR dataset is presented in Table~\ref{table:viperall}. Both NP-MFML and SM-MFML have competitive performance against all the other metric learning methods. It should be  noted that the proposed methods again outperform LFDA and kLFDA with significant margin. This re-emphasize that our approach of metric learning without using any local neighborhood information is more efficient for person re-ID on small datasets with few samples per class. NP-MFML and SM-MFML are also superior to the multiple kernel method MKML \cite{MKML}, indicating that our proposed approaches are efficient in  utilizing the information from multiple kernels. 
\setlength{\parskip}{0em}

The proposed methods attain superior performance even against the deep learning methods. This signifies the limitation of the deep learning methods for person re-ID, especially on small size dataset. Our proposed methods also attains superior performance against most of the re-ranking methods including SSDAL, MuDeep, OL-MANS, DLPAR and PDC. In Fig. \ref{fig:teaser} (b), we show few sample probe images of VIPeR dataset and their ranking results from the gallery using NP-MFML.\\

\setlength{\parskip}{-0.75em}
\setlength{\parskip}{0em}
\subsection{Analysis of the Proposed Method}

\begin{figure}[t]
\begin{minipage}{1\linewidth}
  \centering
  \centerline{\includegraphics[width=7cm]{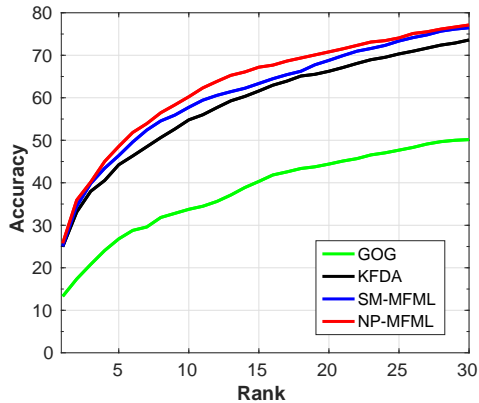}}
\end{minipage}
\caption{Influence of components: Cumulative Matching Curves (CMC) obtained using the GOG feature descriptor alone, KFDA, NP-MFML, and SM-MFML, evaluated using GRID dataset.}
\label{fig:comp}
\end{figure}

\noindent\textbf{Influence of Components}:
In order to analyze the contribution of each of the components in our proposed framework, we separately evaluate their accuracies attained by each component. We use GRID dataset for the evaluation. In Fig. \ref{fig:comp}, we plot Cumulative Matching Curves (CMC) curves obtained using the feature descriptor alone (without metric learning), KFDA (single kernel) and the multiple kernel approaches NP-MFML and SM-MFML. 
CMC curves are obtained by plotting the ranking accuracies against the corresponding ranks.  Using the feature descriptor alone, the rank-1 accuracy attained is 13.28\% only. After applying Mahalanobis metric learning using KFDA, the rank-1 accuracy increases to 24.96\%. This clearly demonstrates the significance of metric learning methods in person re-ID. Incorporation of multiple kernels in the metric learning using our proposed frameworks NP-MFML and SM-MFML further increase the rank-1 accuracy to 25.76\% and 25.04\% respectively, illustrating that both NP-MFML and SM-MFML are able to utilize the information from multiple kernels to efficiently boost the accuracy.\\

\begin{figure}[t]
\begin{minipage}{1\linewidth}
  \centering
  \centerline{\includegraphics[width=7cm]{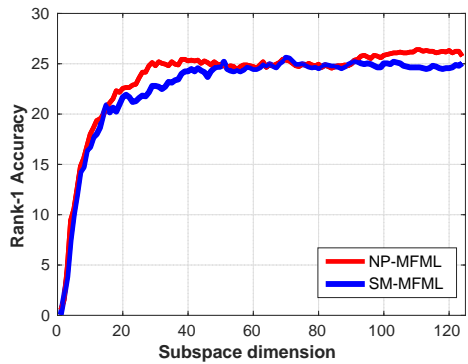}}
\end{minipage}
\caption{ Influence of subspace dimension of the proposed methods on their rank-1 accuracies, evaluated using GRID dataset.}
  \label{fig:subs}
\end{figure}

\setlength{\parskip}{-0.5em}
\noindent\textbf{Influence of  Subspace Dimension}: We also study the influence of the dimension of the underlying discriminative subspace of the proposed metric learning methods NP-MFML and SM-MFML on the person re-ID accuracy. In Fig. \ref{fig:subs}, we plot the rank-1 accuracy verses the subspace dimension, evaluated using the GRID dataset. The subspace discriminants are sorted based on their corresponding eigenvalues. Initially, the rank-1 accuracy keeps increasing with the number of discriminant vectors (that define the projected subspace dimension) and later becomes almost stable at higher dimensions. For practical computation, we can also choose a much smaller dimension as more than 95\% of the maximum accuracy is already attained using the first 50 discriminants. \\

\noindent\textbf{Runtime Analysis:} We compare the testing time of our proposed metric learning methods with other metric learning methods including  KNFST \cite{Zheng:nfst}, MFA \cite{rPcca}, XQDA \cite{GOG}, MLAPG \cite{MLAPG} and kLFDA \cite{rPcca}, using VIPeR dataset. Our codes for KFDA, NP-MFML and SM-MFML are all implemented using MATLAB. All methods are executed on a PC with Intel i7-6700CPU@3.40GHz and 32GB memory. There are 632 samples in the training set and 316 queries in the test set. As summarized in Table~\ref{executionTime}, our proposed methods have testing time comparable to other methods.\\
\setlength{\parskip}{0em}

\begin{table}[h]
\caption{ Test run-time evaluation (in seconds) on VIPeR dataset.}
\begin{center}
\begin{tabular}{lc}
\hline
Methods  & Time\\
\hline
KNFST& 0.52 \\
MFA & 3.99  \\
XQDA & 0.33 \\
MLAPG & 0.13\\
\hline
\end{tabular}
\hspace{0.5em}%
\begin{tabular}{lc}
\hline
Methods  & Time\\
\hline
kLFDA &4.13 \\
\textbf{KFDA} & 0.52 \\
\textbf{NP-MFML} &  0.57 \\
\textbf{SM-MFML} & 0.56 \\
\hline
\end{tabular}
\label{executionTime}
\end{center}
\end{table}

\section{Conclusions}
\label{sec:Conclusion}
In this paper, we have proposed a Mahalanobis metric learning framework induced from Kernel Fisher Discriminant Analysis for person re-identification. The metric learns an embedding space where the samples of distinct classes are well separated and the samples of the same class come closer. We demonstrated that our KFDA induced distance metric, without using any supplementary technique, performs superior to the methods that learn local neighborhood information, especially in small datasets with very few samples per class. We also proposed to incorporate information from multiple kernels for the metric learning to further enhance the discrimination. The proposed multiple kernel learning frameworks efficiently remove the bias in using a single kernel and improves the accuracy in person re-identification. Our extensive experiments on three benchmark datasets confirm that the performance of the proposed methods are very competitive with  state-of-the-art methods.

\noindent{Acknowledgment}
This research work is supported from Ministry of Electronics and Information Technology (Meity), Government of India, under Visvesvaraya PhD Scheme. We also acknowledge the support provided by National Center of Excellence in Technology for Internal Security (NCETIS) (an initiative by IIT Bombay and Deity).